\documentclass{article}
\usepackage[final,nonatbib]{nips_2016}

\pdfoutput=1

\usepackage{amsthm} 
\usepackage{amssymb}  
\usepackage{amsmath} 
\usepackage{graphicx}
\usepackage{graphics}
\usepackage{enumerate}
\usepackage{txfonts} 

\usepackage[bf]{caption}
\usepackage{subfig}

\newcommand{\E}{{\mathbb{E}}}

\newcommand{\Dxite}{{{D_x}^{k,test}}}

\newcommand{\fhat}{{\hat{f}}}

\newcommand{\ftilde}{{\tilde{f}}}
\newcommand{\ghat}{{\hat{g}}}

\newcommand{\gtilde}{{\tilde{g}}}

\newcommand{\fhatsmc}{{\hat{f}_{StackMC}}}

\begin{document}
\title{Reducing the error of Monte Carlo Algorithms by Learning Control Variates}
\author{Brendan D. Tracey \\ Santa Fe Institute \\ Massachusetts Institute of Technology \And David H. Wolpert \\ Santa Fe Institute \\Massachusetts Institute of Technology \\ Arizona State University}
\maketitle

\begin{abstract}
Monte Carlo (MC) sampling algorithms are an extremely widely-used technique to estimate expectations of functions $f(x)$, especially in high dimensions. Control variates are a very powerful technique to reduce the error of such estimates, but in their conventional form rely on having an accurate approximation of $f$, \emph{a priori}.  Stacked Monte Carlo (StackMC) is a recently introduced technique designed to overcome this limitation by fitting a control variate to the data samples themselves. Done naively, forming a control variate to the data would result in overfitting, typically \emph{worsening} the MC algorithm's performance. StackMC uses in-sample / out-sample techniques to remove this overfitting.  Crucially, it is a post-processing technique, requiring no additional samples, and can be applied to data generated by \emph{any} MC estimator. Our preliminary experiments demonstrated that StackMC improved the estimates of expectations when it was used to post-process samples produces by a ``simple sampling" MC estimator. Here we substantially extend this earlier work. We provide an in-depth analysis of the StackMC algorithm, which we use to construct an improved version of the original algorithm, with lower estimation error. We then perform experiments of StackMC on several additional kinds of MC estimators, demonstrating improved performance  when the samples are generated via importance sampling, Latin-hypercube sampling and quasi-Monte Carlo sampling. We also show how to extend StackMC to combine multiple fitting functions, and how to apply it to discrete input spaces $x$.
\end{abstract}

\section{Introduction}
In many Machine Learning (ML) applications, one needs to compute the expected value of a quantity.
In continuous spaces, this means estimating the value of an integral of the form
\begin{equation}
\label{eq:EV}
\E[f] = \fhat = \int f(x)p(x)\,dx
\end{equation} 
where $f(x)$ is the function of interest, and $p(x)$ is the probability of the input $x$. Often in ML contexts this integral is not known analytically, and so must be estimated from a finite set of samples of the function --- which requires an estimator
algorithm for mapping that set to an estimate of the integral. When the samples are generated deterministically -- as in quadrature -- the quality of the estimator is just its error. When the samples are instead generated randomly -- as
in Monte Carlo (MC) algorithms -- the quality of the estimator can be quantified as its expected squared error. 

One way to reduce the error of an estimator based on MC samples is by exploiting some side information.
An important example of such information is a ``control variate'', which is a function with known mean whose behavior is correlated with $f(x)$. The difference between the sample mean of the control variate
and its actual mean provides information about the difference between the sample mean of $f$ and \emph{its}
true mean -- information which can then be exploited to correct that sample mean to provide a better estimate of the integral. 
Unfortunately, most problems of interest do not have a natural control variate to perform such regularization.
The core idea of the Stacked Monte Carlo (StackMC) algorithm \cite{tracey2013using} is to \emph{construct} a control variate by training a supervised learning algorithm on the available data samples. The supervised learning algorithm is chosen such that the expected value of the fit can be found analytically (or cheaply through sampling). Then, the fit is evaluated on held-out data samples, and the discovered performance is used to estimate the quality of the control variate. The original presentation of StackMC tested the algorithm under simple sampling, finding StackMC has expected error at least as low as the better of MC and the fitting algorithm alone.

Here we provide a deeper theoretical understanding of StackMC and extend StackMC to more sophisticated MC techniques. First, we review the original presentation of StackMC and compare the estimation procedure with other algorithms. We examine some of the implicit assumptions, and find  an improved estimator for the quality of the fit. We then present modifications that extend the algorithm to new regimes of interest, specifically when samples are generated from quasi-Monte Carlo, when samples are generated from importance sampling, and finally when the input domain is discrete instead of continuous. We find that with the appropriate modifications, the estimate under StackMC has error at least as low as the Monte Carlo estimate in almost all cases, and in most cases the StackMC error is also at least as low as the fitting algorithm used alone.

\section{Stacked Monte Carlo}
\label{sec:smcback}
The estimation error of an arbitrary estimator, $E[(\fhat_{est} - \fhat)^2]$, can be decomposed as $v + b^2$, where $v$ is the variance of the estimator, $E[(\fhat_{est} - E[\fhat_{est}])^2]$, and $b$ is the bias of the estimator, $E[\fhat_{est} - \fhat]$. An estimation algorithm that has small expected squared error is one that maintains both a low bias and a low variance. The original presentation of StackMC \cite{tracey2013using} imagines that there exists a function $g$ that is correlated with $f$ and has $\ghat$ as its known mean. This function is combined into \eqref{eq:EV},
\begin{eqnarray}
\label{eq:smc1_nofold}
\hat{f} &= & \int{\alpha g(x)p(x)}\,dx + \int{\big{(}f(x)-\alpha g(x)\big{)}p(x)}\,dx \nonumber \\
\label{eq:intstackmc}
        &= & \hspace*{0.37in} \alpha \hat{g}                          \hspace*{0.37in}+\int{\big{(}f(x)-\alpha g(x)\big{)}p(x)}\,dx
\end{eqnarray}
where $\alpha$ is a constant. The function samples and corresponding $g$ values are used to estimate \eqref{eq:intstackmc},
\begin{equation}
\label{eq:origstackmc}
\fhat_{smc} = \alpha \hat{g} +\frac{1}{N} \sum_i f(x_i) - \alpha g(x_i)
\end{equation}
 At first glance it seems nothing has been accomplished since we have merely replaced one integral equation with another. However, since the control variate $g$ is correlated with $f$ by assumption, if $\alpha$ is chosen properly then $f - \alpha g$ has lower variance than $f$. The estimation error of Monte Carlo is directly proportional to the variance of the function whose mean is being estimated. This new integral has lower variance, and thus lower estimation error. In fact, the stronger the correlation of $g$ is correlated with $f$, the more accurate the MC estimate of $\int (f(x) - \alpha g(x)) p(x) dx$.

The difficulty, of course, is how to find $g$ in the first place. One natural idea is to construct it directly from the pairs $(x, f(x))$ using a supervised learning algorithm that produces fits whose value of $\ghat$ can be found through analytic integration or sampling. Unfortunately this approach is subject to over-fitting, which typically introduces very large bias into the resultant integral estimator. To overcome this over-fitting problem, it is natural to consider using the out-of-sample behavior of the supervised learning algorithm on the pairs $(x, f(x))$ to correct the integral estimate. 

Sub-sampling techniques, such as cross-validation, frequently use out of sample behavior to decide upon a single best fit. In context, this would be choosing a single best $g$ to use in \eqref{eq:origstackmc}. However, just as one might expect a combination of features to produce a better fit than a single feature, one may also expect that a combination of fits to in-sample data would provide a better estimate than any single one of them. This idea of ``blending'' fits is the underlying idea of stacking~\cite{wolp92}. Stacking has a long history of success, being applied to regression \cite{BreimanSR}, classification~\cite{dvzeroski2004combining} probability density estimation \cite{Smyth}, confidence interval estimation \cite{Kim}, and optimization \cite{Dev}. 

StackMC uses the predictions at out-of-sample locations to find the value of $\alpha$ and set $g(x_i)$. In particular, StackMC partitions the data into $K$ sets (as in $K$-fold validation). Each of these sets is the held-out data for a single fold, and the corresponding held-in set is the complement set to the held-in set. The value of $g(x_i)$ is taken to be the value predicted at $x_i$ when $x_i$ is in the held-out data (with $K$-fold ensuring every data location is held-out exactly once). StackMC then uses these held-out predictions and their corresponding true function values $f(x_i)$ to estimate $\alpha$ using the equation from control variates, $\alpha = cov(g,f) / var(g)$. Finally, the expected value for each fold is corrected using the held-out samples, and $\fhat$ is estimated as the average of the fold expected values, where $m_k$ is the number of held-out points in fold $k$.
\begin{equation}
\label{eq:smc1}
\fhatsmc=\frac{1}{K}\sum_{k=1}^K \fhatsmc(i) =\frac{1}{K}\sum_{k=1}^K{\bigg{[}\alpha \hat{g_k}  + \frac{1}{m_k} \sum_{j=1}^{m_k}{f \big{(}\Dxite(j) \big{)}-\alpha g_i \big{(}\Dxite(j) \big{)}}\bigg{]}} \quad \text{.}
\end{equation}

\subsection{Comparison with other methods}


There are a multitude of different Monte Carlo methods, each with its own way of reducing variance. For the purposes of this work, there are two broad classes of MC algorithms: those where sampling from $p$ is easy but evaluating $f$ is difficult, and those where sampling from $p$ directly is difficult or impossible. This later regime contains the Metropolis-Hastings algorithm \cite{zhang2011quasi} and its many variants such as Gibbs sampling and Hamiltonian Monte Carlo \cite{seiler2014positive, neal2011mcmc}. Regularizing $f$ with $g$ does not help if it is $p$ which is hard to acquire (for example, $\ghat$ could not be found), and combining StackMC with these methods is outside the scope of this work.

When $f$ is expensive to evaluate, one wants to be judicious in the number of samples required. Many techniques seek to be smarter in how samples are allocated. For example, importance sampling \cite{mackay1998introduction} generates samples from an alternate distribution $q(x)$ to bias the samples into ``important'' regions. Cross entropy methods \cite{rubinstein2013cross,ravanbakhsh10,szita2006learning} are the adaptive version of importance sampling that try to actively modify $q(x)$ to be close to the ideal importance sampling distribution. Quasi-Monte Carlo methods seek to distribute samples evenly, while others use Bayesian priors \cite{vlassis2012bayesian,ghahramani2002bayesian,gunter2014sampling} to intelligently place sample locations.  In contrast to all of these techniques, StackMC is a post-processing method. StackMC does not change how the samples are generated, and so it can, at least in theory, be combined with any of the sampling methods mentioned above. In fact, several such combinations are demonstrated in this work. StackMC is most closely related to recent developments in control functionals \cite{oates2015control, oates2014control}, though these works focus on constructing good kernel-based fits.

\section{Detailed Analysis and Extensions}
Above we described StackMC as it was initially presented, and contrasted it with other variance reduction methods. This section first provides greater detail on the estimation error of StackMC, focusing on the estimation of $\alpha$. The section then presents modifications to the StackMC algorithm to apply it in sampling regimes beyond simple Monte Carlo. Accompanying this analysis are computational results demonstrating variance reduction. These result plots depict the expected squared error for the StackMC algorithm being tested, for the Monte Carlo sampler, and for the fit alone. The squared error for the fit alone is computed by training the supervised learning algorithm on all of the available data samples and estimating $\fhat$ as $\ghat$. The error bars are the error in the mean.

\subsection{Estimator Error}
\label{sec:detail}
As described in Sec. \ref{sec:smcback}, StackMC estimates the value of $\fhat$ as an average of the predictions from the held-out folds. Using \eqref{eq:smc1}, it can be seen that the expected squared error of this estimator is
\begin{align}
\label{eq:smcee}
E[(\fhatsmc - \fhat)^2] &= E \left[ \left( \frac{1}{K} \sum_k \alpha \ghat_k + \frac{1}{N} \sum_i f_i  - \alpha g_i - \fhat \right)^2\right]
\end{align}
where the expectation is over datasets, $\ghat_k$ is the expected value from the held-in data in fold $k$, and $g_i$ is the prediction made by the fitter to the fold for which $x_i$ is in the held-out dataset. The sum over $j$ and $k$ simplifies to the single sum over $i$ above under $K$-fold validation since $m_k$ is constant across $k$ and each $x_i$ is held-out exactly once. The above can also be written as
\begin{equation}
E[(\fhatsmc - \fhat)^2] = E\left[ \left( \left( \frac{1}{N} \sum_i f_i - \fhat \right) - \alpha \left( \frac{1}{N} \sum_i g_i - \frac{1}{K} \ghat \right) \right)^2 \right]
\end{equation}
For simplicity, make the following substitutions: $\gtilde = 1/N \sum_i g_i$, $\ghat = 1/K \sum_k \ghat_k$ and $\ftilde$ for $\frac{1}{N} \sum_i f_i$.
\begin{align}
&= E\left[ \left(\left( \ftilde - \fhat \right)  - \alpha \left( \gtilde - \ghat \right) \right)^2 \right] = E \left[ (\ftilde - \fhat)^2 - 2 \alpha (\gtilde - \ghat)(\ftilde -\fhat) + \alpha^2 (\gtilde - \ghat)^2  \right] \\
\label{eq:smcprealpha}
&= E \left[ (\ftilde - \fhat)^2 \right]  -2 \alpha E\left[ \left(\gtilde - \ghat \right) \left(\ftilde -\fhat \right) \right] + \alpha^2 E\left[ \left(\gtilde - \ghat \right)^2 \right]
\end{align}
The first term in \eqref{eq:smcprealpha}, $E \left[ (\ftilde - \fhat)^2 \right] $, is the error under the sampling procedure. If $\alpha$ is chosen such that 
\begin{equation}
2 \alpha E\left[ \left(\gtilde - \ghat \right) \left(\ftilde -\fhat \right) \right]  > \alpha^2 E\left[ \left(\gtilde - \ghat \right)^2 \right]
\end{equation}
then the error of this estimation procedure is lower than the original Monte Carlo estimate. 

This illuminates the importance of choosing $\alpha$ properly. It is not the case that setting $\alpha$ to a fixed constant, say $\alpha = 0.5$, necessarily leads to variance reduction. In fact, a value of $\alpha$ that is too large will actually \emph{increase} the variance of the estimator. The value for $\alpha$ which minimizes the variance is found by taking the derivative with respect to alpha and setting it to zero. The second derivative is always positive, so this estimator for $\alpha$ is the unique global minimizer of $\alpha$:
\begin{align}
\label{eq:goodalpha}
\alpha^* = \frac{E\left[ \left(\gtilde - \ghat \right) \left(\ftilde -\fhat \right) \right]}{E\left[ \left(\gtilde - \ghat \right)^2 \right]} = \frac{cov(\gtilde - \ghat, \ftilde ) + b_g b_f}{var(\gtilde - \ghat) + {b_g}^2}
\end{align}
where $b_g = E[\gtilde - \ghat]$ and $b_f = E[\ftilde - \fhat]$. 

This optimal value for $\alpha$ cannot be estimated from the samples, as there is only data from 1 $K$-fold partition. While one could make several $K$-fold partitions, the estimate of $\ftilde$ is the same for all of them. However, one can approximate this equation by using the per-fold information by taking $\ghat$ as simply the expected value for the fold, and estimate $\gtilde$ and $\ftilde$ using only the held-in samples. Furthermore, it is impossible to directly estimate $b_f$ since of course $\fhat$ is unknown, but many sampling distributions are known to be unbiased, and so $b_f = 0$. With these approximations, the optimal estimator for $\alpha$ becomes 
\begin{equation}
\label{eq:newalpha}
\alpha^* = \frac{cov(\gtilde - \ghat, \ftilde)}{var(\gtilde - \ghat) + {b_g}^2}
\end{equation}

This equation for $\alpha^*$ is, in general, different than $\alpha^* = cov(g_i,f_i) / cov(g_i)$, the estimator used in     \cite{tracey2013using}.  These two estimates become the same when: 1) $K = N$  (leave-one-out), 2) $b_g = 0$, and 3) $\ghat$ is constant. The first assumption is reasonable, as it gives $N$ data points with which to compute $\alpha$ instead of only $K$. These latter assumptions characterize the difference between the two estimators. If the held-out samples are not an unbiased estimator of the mean of the fit, then $b_g \ne 0$. (This case is discussed further below.) When $b_g = 0$, this third assumption is similar to the statement  $cov(\gtilde, \ftilde) >> cov(\ghat, \ftilde)$, i.e., the covariation of the fit sample mean is much larger than the covariation of the mean of the fit. This is a reasonable assumption when the number of samples is large, since the fit should stabilize as the number of samples grows large. However, if the sample
size is small, this assumption may not hold. Performance could be improved by using \eqref{eq:newalpha} instead of the original estimator, or perhaps with the additional regularization of setting $b_g = 0$ if it is known to be true.

This equation for $\alpha$ can be inserted back into \eqref{eq:smcprealpha} to find the expected error reduction given optimal estimation for $\alpha$. Performing this substitution one finds
\begin{align}
\label{eq:smcalpharight}
E[(\fhatsmc - \fhat)^2] &= E[(\ftilde - \fhat)^2] -  \frac{E \left[ \left( \gtilde - \ghat \right) \left( \ftilde - \fhat \right) \right]^2}{E \left[ \left( \gtilde - \ghat \right)^2 \right]}
\end{align}
Having the optimal value of $\alpha$ guarantees a reduction in expected squared error, as the Monte Carlo error is reduced by a strictly positive term. This equation is similar to the standard control variates result, but, again, differs in that the individual estimators $\ftilde$ and $\gtilde$ may be biased, and $\ghat$ may fluctuate. If $b_f = b_g = 0$, and $\ghat = c$, then the standard control variates formula is recovered:
\begin{equation}
E[(\fhatsmc - \fhat)^2]  = \left( 1 - \rho^2  \right) var(\ftilde)
\end{equation}
where $\rho$ is the correlation between $\gtilde$ and $\ftilde$. This highlights that a good fitting algorithm will produce a high correlation between $\ftilde$ and $\gtilde$, while keeping the variation in the mean across fits small.

\subsubsection{Updated Alpha Estimation Example}
\begin{figure*}
\centering
\begin{minipage}[b]{.49\textwidth}
\includegraphics[width=1.0\textwidth]{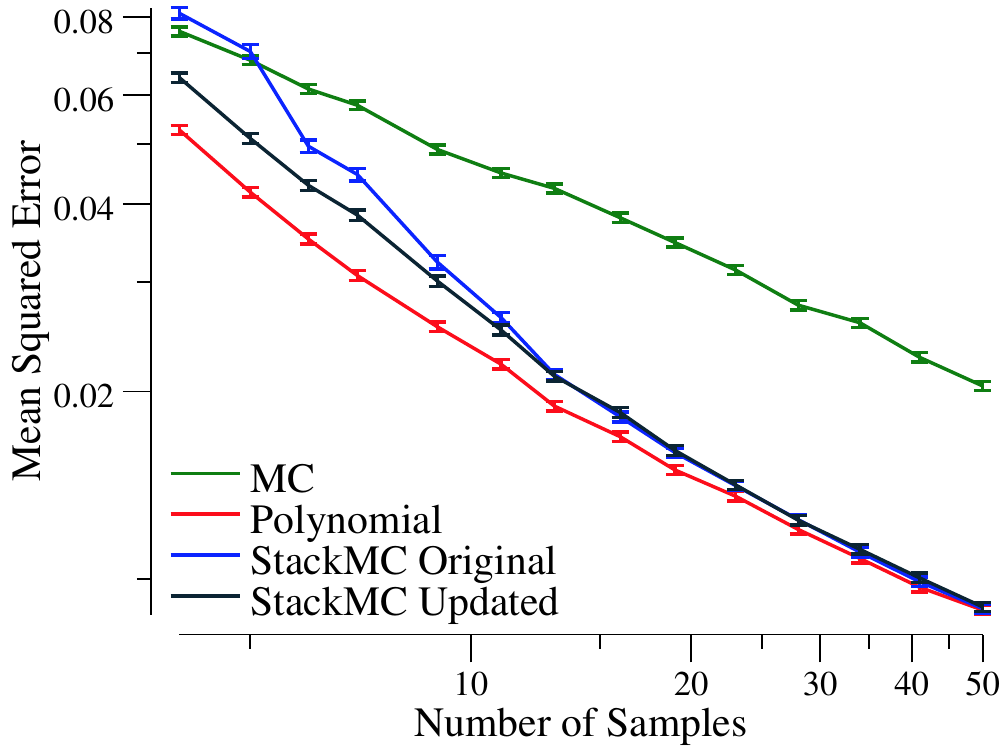}
\caption{Comparison of the squared error using the original StackMC alpha estimate and the updated alpha estimate. There is a noticeable improvement in performance at small sample sizes}\label{fig:UpdatedAlpha}
\end{minipage}\hfill
\begin{minipage}[b]{.49\textwidth}
\includegraphics[width=1.0\textwidth]{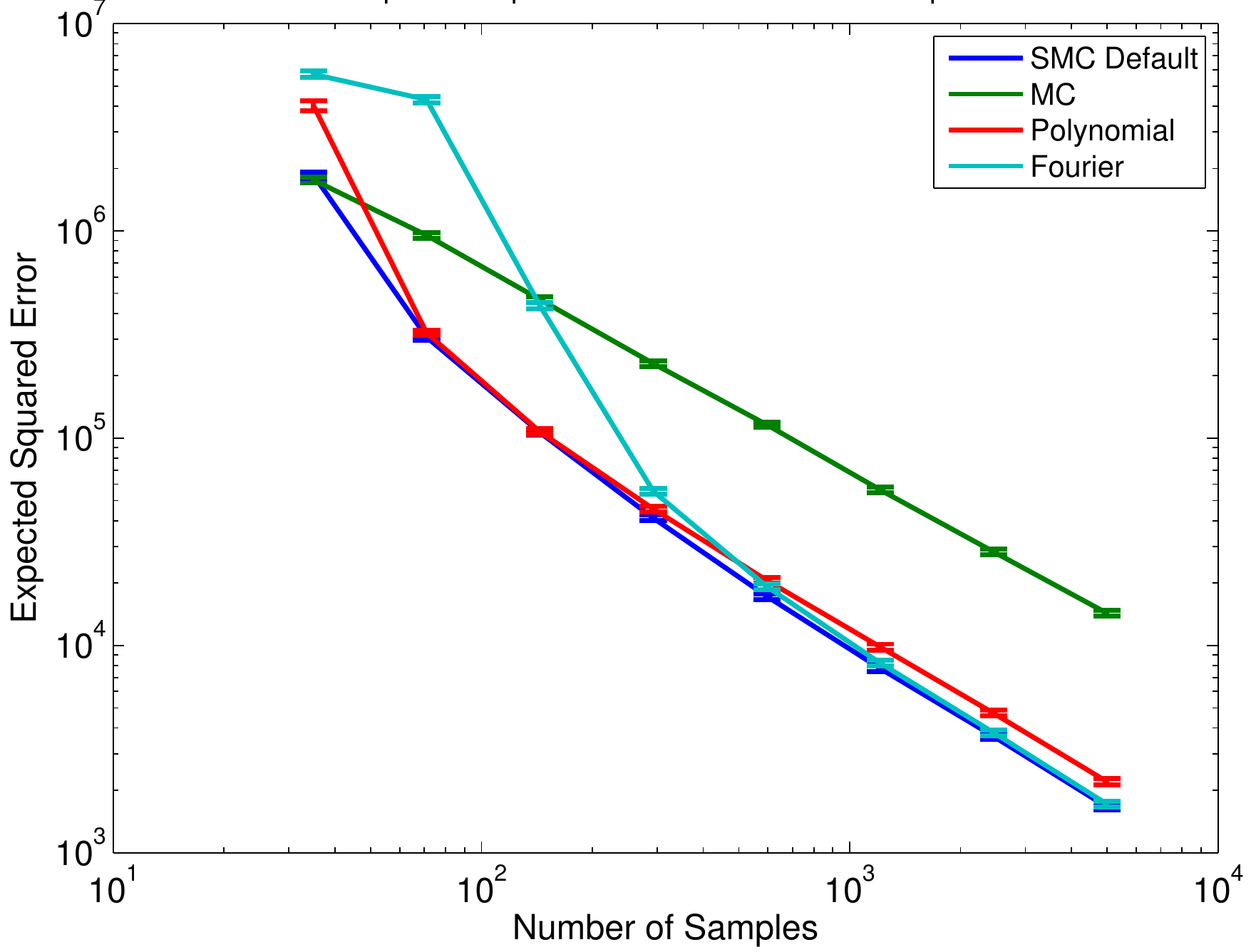}
\caption{Expected squared error comparison for multiple supervised learning algorithms. StackMC has a squared error equal to the lowest of any of the algorithms.}\label{fig:RosenUnifPolyFourier10D}
\end{minipage}
\end{figure*}

As mentioned, this new estimate of $\alpha^*$ is likely to be better when there is large variation of $\ghat$ across the folds. This is likely to happen when the sample size is only slightly larger than the number of free parameters. We test this hypothesis by constructing a simple example. The function of interest is a simple quadratic, $f(x) = (x - 0.2)^2$, and the sampling distribution is uniform over the unit interval. The fitting algorithm is a linear fit to the held-in data samples. The results of this test case are shown in Fig.\ref{fig:UpdatedAlpha}, and they confirm the hypothesis. At very small numbers of sample points, there is significant improvement by including the fluctuations of $\ghat$ in the estimate of $\alpha$. As the number of samples grows, this effect becomes negligible, and the two estimators have equivalent performance. 

\subsection{Multiple Fitting Algorithms}
Frequently, one may consider multiple different fitting algorithms to train on the data. Instead of choosing among them, they may all be used together to jointly reduce the estimation error. Given a set of supervised learning algorithms, \eqref{eq:EV} can be written as 
\begin{equation}
\fhat = \alpha_1 \hat{g}^{(1)} + \alpha_2 \hat{g}^{(2)} + ... + \int{\big{(}f(x)-\alpha_1 g^{(1)}(x)\big{)}p(x)}\,dx + \int{\big{(}f(x)-\alpha_2 g^{(2)}(x)\big{)}p(x)}\,dx + ... \quad \text{.} 
\end{equation}
which introduces some number of control variates, $\hat{g}^{(i)}$. Similar to the analysis above, the expected error of this estimator is 
\begin{align}
\label{eq:multismc}
E[(\ftilde -\fhat)^2] - 2 \sum_i \alpha_i  E[(\gtilde_i - \ghat_i)(\ftilde - \fhat)] + \sum_i \sum_j \alpha_i \alpha_j E[(\gtilde_i - \ghat_i)(\gtilde_j - \ghat_j)]
\end{align}
Let $u_i  = E[(\gtilde_i - \ghat_i)(\ftilde_i - \fhat)]$ and $W_{i,j} = E[(\gtilde_i - \ghat_i)(\gtilde_j - \ghat_j)]$, so that \eqref{eq:multismc} becomes 
\begin{align}
E[(\fhatsmc - \fhat)^2] = E[(\ftilde -\fhat)^2] - 2 u^T \boldsymbol{\alpha} + \boldsymbol{\alpha}^T W \boldsymbol{\alpha}
\end{align}
This is quadratic in $\boldsymbol{\alpha}$, with minimizer $\boldsymbol{\alpha} = W^{-1} u$. The estimation error using all of the fits is  
\begin{align}
E[(\fhatsmc - \fhat)^2] &=  E[(\ftilde -\fhat)^2]  - 2 u^T W^{-1} u + (W^{-1} u)^T W (W^{-1} u) \\
&= E[(\ftilde -\fhat)^2] - u^T W^{-1} u
\end{align}

The total error reduction from the joint set of fitting algorithms depends on how correlated the fits are with one another and the true function samples. Consider the case where all fitters have equal covariance with the function samples, $u_i = c_u$, and equal variances, $W_{i,i} = c_w \forall i$. Further, assume the off-diagonal terms are represented by $W_{i,j} = \rho_{i,j} c_w$. In the ideal case, the fitters are uncorrelated, $\rho_{i,j} = 0$ and the total variance reduction is $N_{fit} * c_u / c_w$.  On the other hand, if all of the fitters are perfectly correlated with one another, $\rho_{i,j} = 1$, and the error reduction is just $c_u / c_w$, i.e., there is no extra variance reduction from incorporating multiple fits. This shows that a set of fitting algorithms should be sought that each have strong coupling with the true function but weak coupling amongst themselves. In theory, adding a new fitting algorithm should always be beneficial (or at least not harmful), though in practice errors in the estimation of $\alpha$ could lead to degraded performance.

\subsubsection{Multiple Fitting Algorithms Example}
As a test for multiple fitting algorithms, we chose the Rosenbrock function
$f(x) = \sum_{i=1}^{d-1} \left[  (1-x_i)^2+ 100 (x_{i+1} - x_i^2 )^2 \right]$. The Rosenbrock is widely used as a benchmark problem in optimization. In addition, it has features representative of many engineering problems, in that there is a shallow region of good performance with interesting structure, and performance degrades significantly outside that region. The uncertainty was assumed to be a uniform distribution over the [-3,3] hypercube. Two different supervised learning algorithms were used. First, a third-order polynomial with no cross terms, i.e. $g(x)= \beta_0+\sum_{i=1}^d \beta_{1,i} x_i+\beta_{2,i} x_i^2+ \beta_{3,i} x_i^3$, and a Fourier fitting algorithm $g^{(2)}(x)= \beta_0+
\sum_{i=1}^d \sum_{j=1}^6 \beta_{j,1,i} \cos(j*(x_i - x_{min})/(x_{max} - x_{min}) - \pi)+ \beta_{j,2,i} \sin(j*(x_i - x_{min})/(x_{max} - x_{min}) - \pi)$. For both algorithms, the $\beta$ parameters of this polynomial are trained using least squares. The data is in a 10-D space, and the data is partitioned into $5$-folds. The estimation error as a function of number of samples can be seen in Fig.\ref{fig:RosenUnifPolyFourier10D}.

For small sample sizes, the fitting algorithms do quite poorly and the StackMC error equals the MC error. For intermediate sample sizes, the polynomial fit outperforms the other two, but its error is matched by StackMC. Finally, at large sample sizes, the Fourier fit is the best performer, and its error is also matched by StackMC. For all sample sizes, the expected squared error of StackMC is at least as low as the best individually performing algorithm. While it is true that StackMC does not decrease the error for any specific sample size, the best error is obtained without having to choose the best performing algorithm in advance.

\subsection{Quasi-Monte Carlo}
Quasi-Monte Carlo techniques, such as Latin-hypercube sampling and the Halton sequence \cite{Halton}, seek to reduce the variance of Monte Carlo by explicitly spreading the sample locations evenly throughout the space. This sampling procedure is effective at reducing variance, but it also causes all of the data samples to be correlated with one another. In particular, imagine a particular fold $g$ trained on a specific set of held-in data. Under IID sampling, the held-out samples still have a uniform distribution over the domain. However, under Latin-hypercube, conditioned on knowing the held-in samples, the held-out samples have zero probability of being in large regions of the domain. These anti-correlations cause $\alpha$ to be estimated poorly, and thus cause high error in the StackMC estimate. This effect is due to the fact that under quasi-MC, held-out data samples are at regular distances from held-in samples. The covariance between $g$ and $f$ is then typically measured at medium distances from the held-in data. In contrast, under IID sampling, the held-out samples are sometimes nearby to held-in points, and sometimes very far away, netting an overall unbiasedness in cov($g$, $f$). Furthermore,  terms such as  $E[\ftilde - \fhat]$, which are zero under IID sampling, are non-zero on a per-fold basis with quasi-Monte Carlo sampling.

This problem can be mitigated by removing the correlation between held-in and held-out samples. To keep the expected error small, this must be done without introducing significant variance. One algorithm that meets these criteria 
first partitions the data into $k$ folds, and then replaces the held-in samples by doing a bootstap without replacement from the full dataset. This removes much of the correlation between the held-in and held-out samples. In theory, one may be able to reduce correlations further by performing a bootstrap with replacement, but in practice this can cause numerical issues when too many of the same sample are in the held-in set. In practice, this estimator still has relatively high variance due to the in-sample procedure. This can be mitigated by performing this partition/bootstrap procedure several times and setting $\fhat_{smc}$ as the average over all of the runs.

\subsubsection{Quasi-Monte Carlo Example}
\begin{figure*}
\centering
\begin{minipage}[b]{.49\textwidth}
\includegraphics[width=1.0\textwidth]{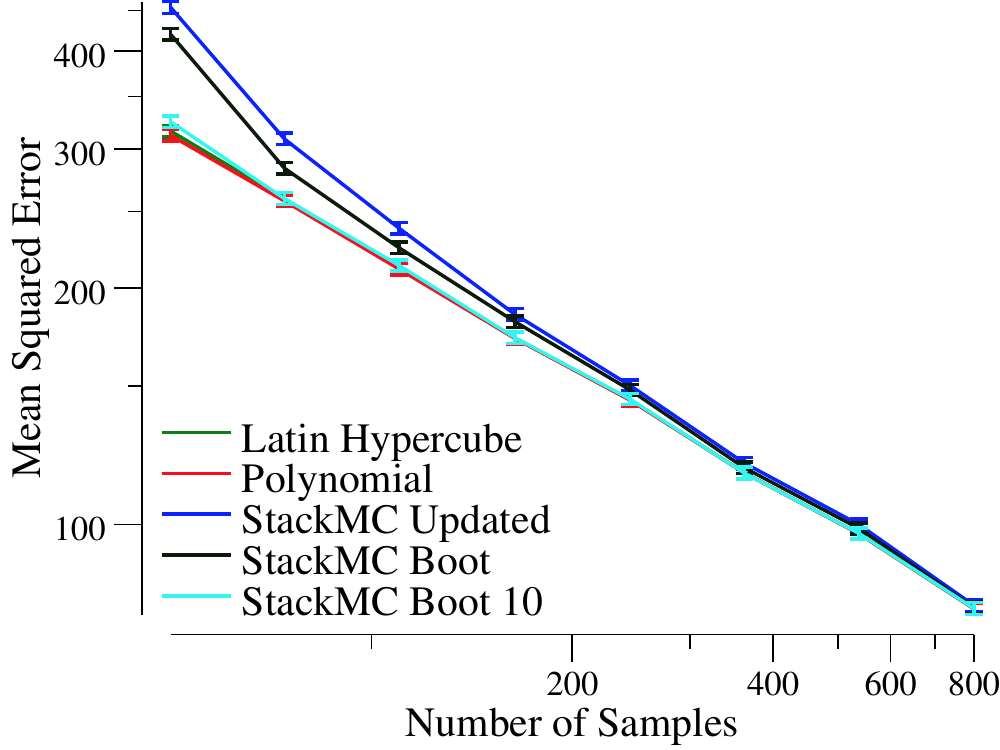}
\caption{Expected squared error comparison under Latin-hypercube generated samples. The standard StackMC estimate has high error due to the sample correlations. The suggested bootstrap procedure reduces this error. }\label{fig:rosenuniflatin}
\end{minipage}\hfill
\begin{minipage}[b]{.49\textwidth}
\includegraphics[width=1.0\textwidth]{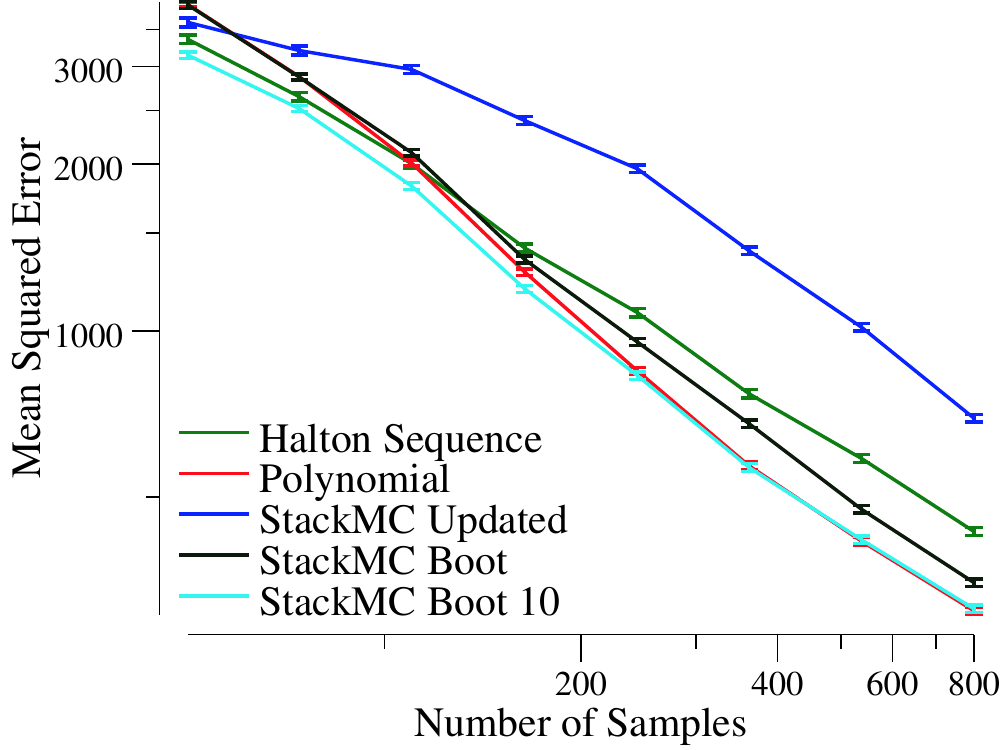}
\caption{Expected squared error comparison under Halton-sequence generated samples. The standard StackMC estimate has high error due to the sample correlations. The suggested bootstrap procedure reduces this error.}\label{fig:rosengausshalton}
\end{minipage}
\end{figure*}

We tested this bootstrap procedure on two different cases. In both $f(x)$ is 10-D Rosenbrock. In the first, samples are generated with Latin-hypercube sampling and a uniform distribution on [-3, 3] in each dimension. In the second case data drawn from the scrambled Halton sequence according to a Gaussian with $\mu = 0, \sigma = 2$ in every dimension. Different draws from the Halton sequence were done with an initial random burn-in length.  The results are seen in Fig.\ref{fig:rosenuniflatin} and Fig.\ref{fig:rosengausshalton}.

In both of these example cases, it is seen that the normal StackMC algorithm has much higher expected error than the Monte Carlo samples. In Fig. \ref{fig:rosengausshalton}, in fact, we see that this performance gap persists even as the number of samples grows. In both of the examples, this bootstrap procedure reduces the estimation error, and when $10$ different samplings are combined, we see that an error at least as low as Monte Carlo is recovered.  For certain regimes in Fig. \ref{fig:rosengausshalton}, the error is in fact lower than the Monte Carlo error and the direct fit error.

\subsection{Importance Sampling} 
\begin{figure*}
\centering
\begin{minipage}[b]{.49\textwidth}
\includegraphics[width=1.0\textwidth]{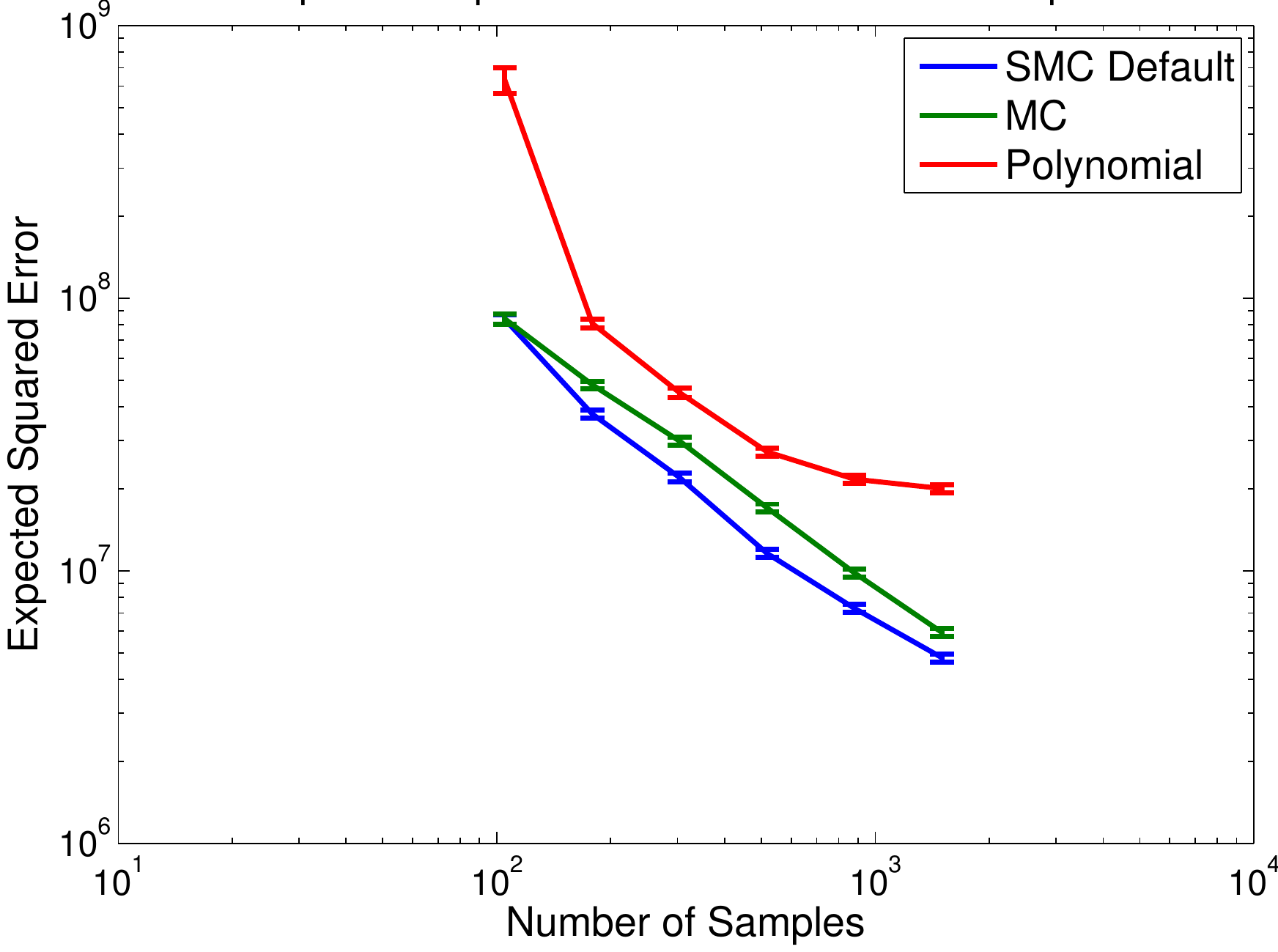}
\caption{Squared error comparison for Latin-hypercube generated samples and $\ghat$ estimated from MC. The error of the fit is high, but StackMC still reduces the squared error.}\label{fig:RosenUnif_ImpSampQuad30D}
\end{minipage}
\begin{minipage}[b]{.49\textwidth}
\includegraphics[width=1.0\textwidth]{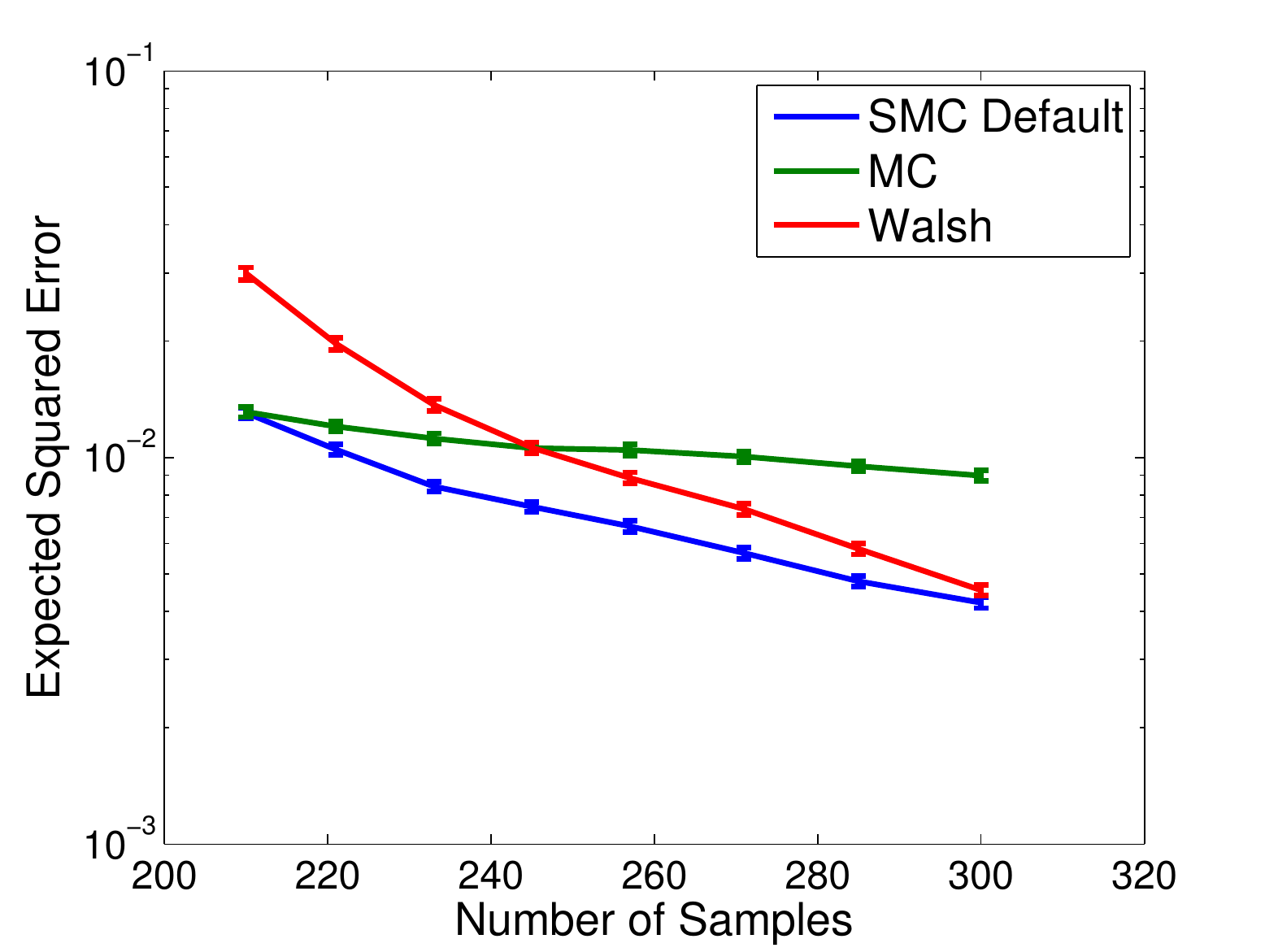}
\caption{Expected squared error comparison using the Walsh learning algorithm in a categorical space. Significant reductions in expected squared error are seen.}\label{fig:TwinPeaks_Walsh}
\end{minipage}\hfill
\end{figure*}

If the sample data are generated via importance sampling, the original StackMC equation no longer holds. The control variates equation is re-written as

\begin{equation}
\hat{f} =  \alpha \int{g(x)q(x)dx} + \int{\Big{(}\frac{f(x)p(x)}{q(x)}-\alpha g(x)\Big{)}q(x)}\,dx
\end{equation}
where $q(x)$ is the importance sampling distribution. The natural choice for the target of the supervised learning algorithm is no longer the true function values themselves, but instead the function values scaled by the importance sampling correction factor $p/q$. 
Like above, we consider the 10-Rosenbrock and a uniform distribution, but this time the samples are generated under the importance sampling distribution  $q(x) = \prod_{i = 1}^d \frac{12}{7} \frac{1}{x_{max} - x_{min}}[(x_i - x_{min})^2/(x_{max} - x_{min}) - 0.5]$. This distribution grows towards the edge of the hypercube like the Rosenbrock function itself. In high dimensions, however, the polynomial fit cannot be integrated analytically under the distribution as there are an exponential number of terms in the product $g(x) q(x)$. The fit must be estimated using Monte Carlo instead. This estimate is performed using $300$ samples generated from $q(x)$, representing a case where samples of $g(x)$ are moderately expensive to produce. The results from this experiment can be seen in Fig. \ref{fig:RosenUnif_ImpSampQuad30D}. It is seen that this high sampling error causes the fit estimation error to be quite high, and yet in the medium range of samples StackMC has lower error than MC. In addition to demonstrating IS-StackMC, this also highlights that $\ghat$ does not need to be estimated exactly.



\subsection{Categorical spaces}
In some cases, one wants to use MC to estimate a sum instead of an integral, for example when $x$ is a boolean string, $\mathbb{B} \equiv \{-1, 1\}$. A fitting algorithm for this space is found by noting that any function of a $d$-dimensional bit string can be represented as 
\begin{equation}
f(x) = \beta_0 +\sum_{i=1}^d \beta_{1,i} x_i + \sum_{i=1}^d \sum_{j=1}^d \beta_{2,j} x_i x_j + \sum_{i=1}^d \sum_{j=2}^d \sum_{k=3}^d x_i x_j x_k + ...  
\end{equation}
This is essentially a discrete Fourier transform of $f$ using an orthonormal basis of the space $(\mathbb{B}^d)^{\mathbb{R}}$ given by the Walsh functions $\{x_1, x_2, \ldots, x_N, x_1 x_2, x_1 x_3, \ldots\}$. An approximation to $f(x)$ can be created by truncating this expansion and only considering contributions from a subset of the Walsh functions. This learning algorithm is a fast ``off the shelf" learner of functions from bit strings to the reals. As an example, we take $f(x)$ to be the the Four Peaks function \cite{baluja1995removing} of Baluja and Caruana, and $p(x)$ to be uniform. The fit ignores all terms with three or more components, and sets the rest using least-squares. This fitting algorithm has very nice property that its expected value is simply $\beta_0$ as all of the other terms have 0 expectation. The results from this experiment are shown in Fig. \ref{fig:TwinPeaks_Walsh}. We see that not only does StackMC have the the lowest squared error for all sample sizes, but for a range of sample sizes the error is much lower than either MC or the Walsh fit on their own.


\section{Conclusions and future work}
StackMC uses supervised learning to construct a control variate from MC samples and then uses stacking to reduce the resultant overfitting problem. It is a purely post-processing technique, applicable to any MC estimator, and can significantly reduce the variance of Monte Carlo integral estimators without adding bias, thereby significantly improving accuracy. 

Here we derive expressions for StackMC's expected error, and use it to
motivate a new estimator of StackMC's parameter
$\alpha$, which can reduce StackMC's error in extreme cases. We also extend StackMC to incorporate multiple control variates; to use data generated with importance sampling and with quasi-Monte Carlo; and to categorical sample spaces. We present experiments verifying the power of these extensions: 
applying them to data generated by an MC algorithm never results in higher error than the 
uncorrected MC estimate, never results in higher error than the fitter (except for very small data sets), and frequently significantly outperforms both. We also find that StackMC may be more flexible than previously appreciated; our results suggest that obtaining an accurate estimate of $\ghat$, the integral of the fit, may not be necessary just to improve over the original MC estimate. 

There are several major areas of future research. We intend to investigate the use of StackMC
for MCMC methods. (MCMC methods create correlations among the samples similar to those
of quasi-Monte Carlo, and so we may need to adapt StackMC for MCMC similarly to how we
did it for quasi-MC.) We also intend to extend StackMC to cases where the probability distribution defining the
desired expectation value is unknown. (In theory, a control variate could be used to estimate the probability distribution itself, and StackMC style techniques applied with that control variate.) 

\end{document}